\documentclass[a4paper,11pt,dvipsnames]{article}

\newif\ifanon
\anontrue

\usepackage{ismo}

\newcommand\cites[1]{\citeauthor{#1}'s\ (\citeyear{#1})}

\usepackage{xcolor}
\usepackage{tikz}

\begin{document}




\ismotitle{The role of attraction-repulsion dynamics in simulating the emergence of inflectional class systems}

\ismoauthor{Erich R. Round & Sacha Beniamine & Louise Esher \\
   \aff{Surrey Morphology Group; Univ. of Queensland;} & \aff{Surrey Morphology Group} & \aff{CNRS LLACAN} \\
   \aff{Max Planck Inst. for the Science of Human History} \\
}


\section{Introduction}\label{sec:introduction}

Dynamic models of paradigm change can elucidate how the simplest of processes may lead to unexpected outcomes, and thereby can reveal new potential explanations for observed linguistic phenomena. \citet{AckermanMalouf2015} present a model in which inflectional systems reduce in disorder through the action of an attraction-only dynamic, in which lexemes only ever grow more similar to one another over time. Here we emphasise that: (1) Attraction-only models cannot evolve the \textit{\textbf{structured diversity}} which characterises true inflectional systems, because they inevitably remove all variation; and (2) Models with both attraction and repulsion enable the emergence of systems that are strikingly reminiscent of morphomic structure such as inflection classes. Thus, just one small ingredient — change based on dissimilarity — separates models that tend inexorably to uniformity, and which therefore are implausible for inflectional morphology, from those which evolve stable, morphome-like structure. These models have the potential to alter how we attempt to account for morphological complexity.

\section{Structure in inflectional systems}\label{sec:structure}

Inflectional classes \citep[‘rhizomorphomes’,][]{Round2015} constitute groups of lexemes which share inflectional exponents; they are a type of ‘morphomic’, morphology-internal structure, mediating the mapping between content and form in inflection. Within complex inflectional systems in natural language, such structures are common, and are demonstrably both productive and psychologically real for speakers \citep{Enger2014,Maiden2018}; they are claimed to limit the complexity of the inflectional system by offering a systematic, recurrent and predictable means of distributing exponents \citep[cf.][]{Carstairs-McCarthy2010,Blevins2016}.

A matter of ongoing debate is what kind of dynamics could potentially lead to such structure \citep{Maiden2018,Carstairs-McCarthy2010}. In this paper, we use computational iterated learning models to reveal for the first time some of the simplest conditions under which stable inflectional class systems can emerge. The insights afforded are of value both for our theoretical understanding of morphomic structure and for the formulation of explicit mathematical models of paradigm evolution, essential to tasks such as robust quantitative historical inference \citep{KellyNicholls2017}.

\section{Attraction-only models cannot evolve stable inflection classes}\label{sec:attraction-only-issues}

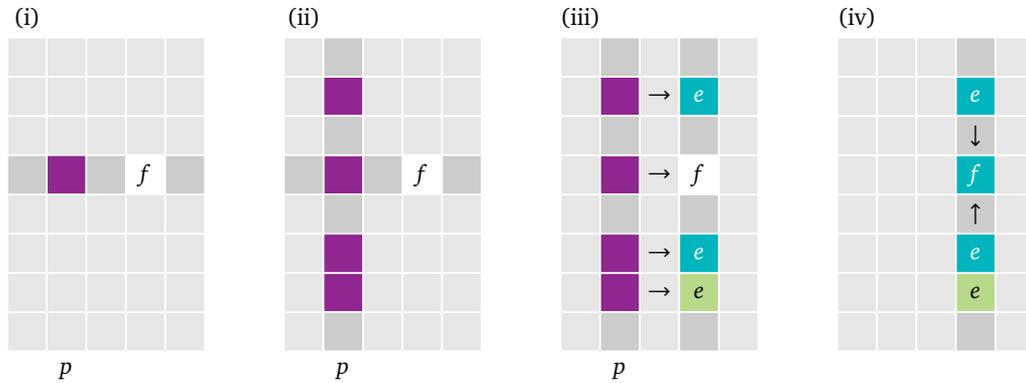
\begin{figure}[htbp]
    \centering
    \resizebox{.85\textwidth}{!}{
\begin{tikzpicture}
    [
        box/.style={rectangle,draw=white,fill=gray!20,very thick, minimum size=1cm},
    ]

\node[draw=none,fill=white] at (0,8){\LARGE(i)};
\node[draw=none,fill=white] at (7,8){\LARGE(ii)};
\node[draw=none,fill=white] at (14,8){\LARGE(iii)}; 
\node[draw=none,fill=white] at (21,8){\LARGE(iv)}; 

\foreach \x in {0,1,...,4}{
    \foreach \y in {0,1,...,7}
        \node[box] at (\x,\y){};
}

\node[box,fill=white] at (3,4){\LARGE\textit{f}}; 
\node[draw=none,fill=white] at (1,-1){\LARGE\textit{p}}; 
\node[box,fill=Plum,text=white] at (1,4){}; 
\node[box,fill=gray!40] at (0,4){}; 
\node[box,fill=gray!40] at (2,4){}; 
\node[box,fill=gray!40] at (4,4){}; 

\foreach \x in {7,8,...,11}{
    \foreach \y in {0,1,...,7}
        \node[box] at (\x,\y){};
}

\node[box,fill=white] at (10,4){\LARGE\textit{f}}; 
\node[draw=none,fill=white] at (8,-1){\LARGE\textit{p}}; 
\node[box,fill=Plum,text=white] at (8,4){}; 

\node[box,fill=gray!40] at (7,4){}; 
\node[box,fill=gray!40] at (9,4){}; 
\node[box,fill=gray!40] at (11,4){}; 

\node[box,fill=Plum,text=white] at (8,1){}; 
\node[box,fill=Plum,text=white] at (8,2){};  
\node[box,fill=Plum,text=white] at (8,6){}; 

\node[box,fill=gray!40] at (8,0){}; 
\node[box,fill=gray!40] at (8,3){}; 
\node[box,fill=gray!40] at (8,5){}; 
\node[box,fill=gray!40] at (8,7){}; 

\foreach \x in {14,15,...,18}{
    \foreach \y in {0,1,...,7}
        \node[box] at (\x,\y){};
}
 
\node[draw=none,fill=white] at (15,-1){\LARGE\textit{p}}; 
\node[box,fill=Plum,text=white] at (15,4){}; 
\node[box,fill=Plum,text=white] at (15,1){}; 
\node[box,fill=Plum,text=white] at (15,2){}; 
\node[box,fill=Plum,text=white] at (15,6){}; 
\node[draw=none,fill=none] at (16,4){\LARGE$\to$}; 
\node[draw=none,fill=none] at (16,1){\LARGE$\to$}; 
\node[draw=none,fill=none] at (16,2){\LARGE$\to$}; 
\node[draw=none,fill=none] at (16,6){\LARGE$\to$}; 
\node[box,fill=white] at (17,4){\LARGE\textit{f}}; 
\node[box,fill=LimeGreen!60,text=black] at (17,1){\LARGE\textit{e}}; 
\node[box,fill=Aquamarine,text=white] at (17,2){\LARGE\textit{e}}; 
\node[box,fill=Aquamarine,text=white] at (17,6){\LARGE\textit{e}};

\node[box,fill=gray!40] at (15,0){}; 
\node[box,fill=gray!40] at (15,3){}; 
\node[box,fill=gray!40] at (15,5){}; 
\node[box,fill=gray!40] at (15,7){}; 

\node[box,fill=gray!40] at (17,0){}; 
\node[box,fill=gray!40] at (17,3){}; 
\node[box,fill=gray!40] at (17,5){}; 
\node[box,fill=gray!40] at (17,7){};

\foreach \x in {21,22,...,25}{
    \foreach \y in {0,1,...,7}
        \node[box] at (\x,\y){};
}

\node[box,fill=Aquamarine,text=white] at (24,4){\LARGE\textit{f}}; 
\node[box,fill=LimeGreen!60,text=black] at (24,1){\LARGE\textit{e}}; 
\node[box,fill=Aquamarine,text=white] at (24,2){\LARGE\textit{e}}; 
\node[box,fill=Aquamarine,text=white] at (24,6){\LARGE\textit{e}};

\node[box,fill=gray!40] at (24,0){}; 
\node[box,fill=gray!40] at (24,3){\LARGE$\uparrow$}; 
\node[box,fill=gray!40] at (24,5){\LARGE$\downarrow$}; 
\node[box,fill=gray!40] at (24,7){}; 

\end{tikzpicture}}
    \caption{A cycle of \cites{AckermanMalouf2015} model. We label the focus~$f$, the pivot cell~$p$, and the evidence~$e$.}
    \label{fig:AM}
\end{figure}

An early iterated learning model, implementing a simple paradigm cell filling task \citep{AckermanBlevinsMalouf2009} in which a lexeme can change only by becoming more similar to another, is described in \citet{AckermanMalouf2015}. The initial input to the model consists of a lexicon in which paradigms are populated with randomly distributed exponents. At each cycle, the model must predict a held out value which we term the \emph{focus}, at the intersection of a focal cell and lexeme. To predict the value of the focus, the model (i) picks a non-focal cell, which we call the \textit{pivot}, (ii) selects all lexemes which share the exponent of the focal lexeme in the pivot cell, (iii) observes the exponents of these lexemes in the focal cell (exponents which jointly constitute \textit{evidence}), and (iv) selects the most frequent exponent in the evidence to replace the held-out focus. Figure~\ref{fig:AM} illustrates one cycle. The result of one cycle is the input to the next.

This model is able to remove disorder from the system, because as lexemes change to be more like others, the dynamic is one of preferential attraction towards exponents that are already more frequent than their competitors. This rich-get-richer dynamic ensures that eventually, all lexemes converge on a single class (though discussion in \citet{AckermanMalouf2015} focuses mainly on transitional states of the system just prior to ultimate uniformity). \citet{AckermanMalouf2015} interpret this result as demonstrating the spontaneous emergence of self-organisational principles in morphological systems. We concur that the model exhibits self-organisation, but only of a radically homogenising kind. Here we investigate a family of  minimally different dynamics (section \ref{sec:our-models}); and their potential to generate outcomes which  more closely resemble the morphological complexity of natural languages (section \ref{sec:results-discussion}).

\section{A family of dynamic, iterated learning models}\label{sec:our-models}

As in \cites{AckermanMalouf2015} model, the initial input to our models consists of a lexicon in which paradigms are populated with randomly distributed exponents, and the basic task is again one of paradigm filling: at each cycle, the model must predict the \textit{focus} based on \textit{evidence} from multiple \textit{pivots}, and the model’s prediction is integrated into the lexicon input to the next cycle. All models in the family are similarly abstract and impoverished: the input provides indices for exponents rather than phonological forms; each cycle represents a general change in the system as a whole, with no method for capturing inter-speaker variation; and the models lack disruptive processes.  

Our innovation is to alter the principles by which the paradigm filling task is accomplished, which we implement as modulable parameters to facilitate the controlled observation of their effects. We allow multiple pivot forms, reflecting the well-established observation that prediction based on multiple forms is more reliable than prediction based on pairings of forms or cells \citep{StumpFinkel2013,BonamiBeniamine2016}. In order to replicate the Zipfian frequency distribution observed for inflectional forms in natural language \citep{BlevinsMilinEtAl2016}, we introduce the option of frequency weighting in two ways: sampling foci in inverse proportion to their lexeme frequency, and sampling pivots and evidence  proportionally to their frequency. Furthermore, we allow the process to be influenced by \textit{negative evidence} \citep{voorspoels2015people}, which introduces an evolutionary \textit{repulsion dynamic}, which at the right strength can enable structured diversity to emerge. To do this, rather than looking only at lexemes which have the same exponent as the focal lexeme in the pivot, we also observe lexemes with different exponents, and use these lexemes to provide evidence of exponents which are expected to be different from the focus. The relative proportion of negative and positive evidence can be varied, and the incorporation of at least some negative evidence proves crucial to the emergence of morphomic structure.

\section{Results and implications of the attraction-repulsion dynamic}\label{sec:results-discussion}

\begin{figure}[htbp]
    \centering
    \includegraphics[width=\linewidth]{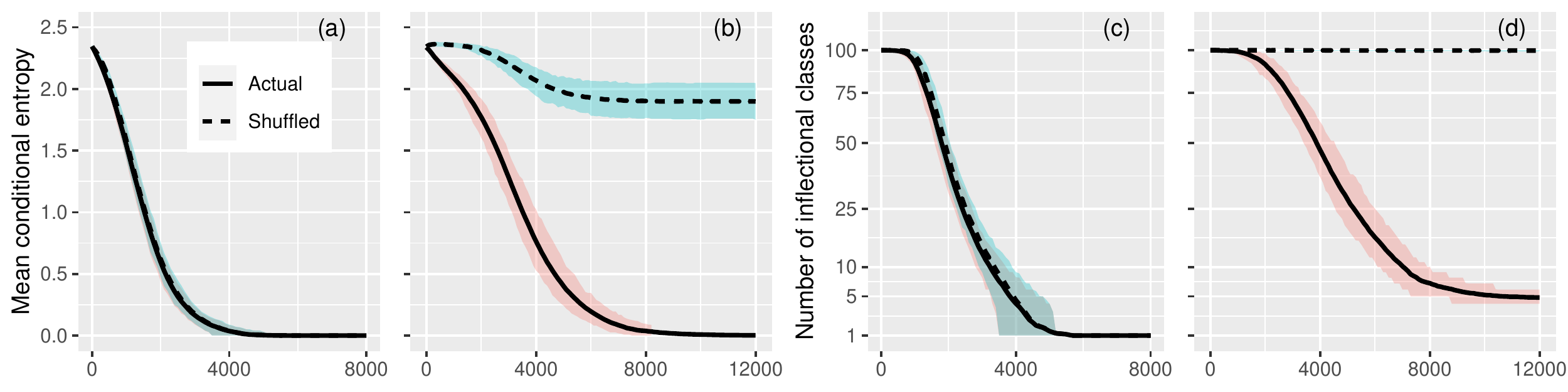}
    \caption{Evolution of mean conditional entropy (a,b) and number of inflectional classes (c,d) for \cites{AckermanMalouf2015} model (a,c) and an attraction-repulsion model (b,d), initial conditions: 100 lexemes, 8 cells, 6 exponents. Lines show means of 100 runs, shading shows 90\% variation. The horizontal axis measures evolutionary cycles.}
    \label{fig:results}
\end{figure}

Figure~\ref{fig:results} compares a replication of \cites{AckermanMalouf2015} model with a model which attends to negative evidence as well as positive (with a 30\%-70\% weighting). Both models were run 100 times and throughout their evolutions we track two measures: firstly, the complexity of the paradigm cell filling task measured as mean conditional entropy between cells \citep{AckermanBlevinsMalouf2009,AckermanMalouf2015}; and secondly, the number of inflection classes (unique rows in the lexicon). As Figure~\ref{fig:results} shows, there is a dramatic reduction in conditional entropy over time in both models. However, we emphasise that this entropy metric has the undesirable property of conflating two kinds of systemic order: (1) order simply due to \textit{\textbf{lack of variation}}, and (2) order due to \textit{\textbf{structured variation}}. Our second metric, the number of inflection classes present, differentiates these two 'orderly' scenarios from one another, and shows that the attraction-repulsion model not only lowers entropy, but also does so while preserving distinct, stable inflectional classes, numbering 4.8 on average. 

We also verify this finding by a second method. If low entropy, or a low number of inflectional classes, is due overwhelmingly to a \textit{lack of variation} at any stage in the evolutionary process, then taking the exponents of any given cell and shuffling them among the language's lexemes should have little effect. This is what we see in the A\&M model (Figure~\ref{fig:results}a,c), where the entropy and number of classes of this `shuffled' version of the evolving system barely differ from the actual system itself. In contrast, if these properties are due to \textit{structured variation}, then shuffling should cause both to be disrupted and both metrics will rise. This is precisely what we observe in the attraction-repulsion model (Figure~\ref{fig:results}b,d).

More generally, we found that a system's progression to uniformity could be slowed by incorporating additional parameters: reducing the amount of available pivots and evidence, or sampling these according to a Zipfian distribution to simulate frequency effects. However, stable inflectional structure still did not emerge in these enriched models, because the dynamic was still one of pure attraction. It is only once negative evidence is attended to, and thus a repulsion dynamic introduced into the system, that the models will consistently develop stable paradigmatic structure. 

\section{Conclusion}\label{sec:conclusion}

We have presented a new and fundamental mechanism for the spontaneous emergence of self-organising structure, which closely resembles what we find in natural language morphologies. Our findings have several topical ramifications. We clarify that morphome-like structure is able to emerge via a dynamic process consisting merely of piecemeal individual changes, within a system which does not explicitly represent morphomic structure (e.g. by means of morphomic indices). The simplicity of the developments involved indicates that, contrary to the prevalent characterisation of morphomic structure as ‘unnatural’ morphology, it is entirely plausible to view such structure as a \emph{natural phenomenon} liable to arise spontaneously within inflectional systems.

\bibliography{ismo}

\end{document}